\documentclass[letterpaper, 10 pt, conference]{ieeeconf}  

\IEEEoverridecommandlockouts                              

\usepackage[dvipsnames,svgnames]{xcolor} 
\definecolor{altncolor}{rgb}{.2,.4,.8} 
\usepackage[colorlinks=true, linkcolor=altncolor, anchorcolor=altncolor, citecolor=altncolor, filecolor=altncolor, menucolor=altncolor, urlcolor=altncolor]{hyperref} 
\usepackage{graphicx}
\usepackage{amsmath,amssymb,amsfonts}
\usepackage{algorithmic}
\usepackage{graphicx}
\usepackage{textcomp}
\usepackage{xcolor}
\usepackage{caption, subcaption}

\usepackage{multirow}
\usepackage{txfonts}

\title{\LARGE \bf
Towards Multi-Object-Tracking with Radar on a Fast Moving Vehicle:\\ On the Potential of Processing Radar in the Frequency Domain 
}

\author{Tim Hansen$^{1}$, Arturo Gomez-Chavez$^{1}$, Ilya Shimchik$^{2}$, and Andreas Birk$^{1}$
\thanks{$^{1}$Authors are with the School of Computer Science and Engineering at Constructor University Bremen, Germany.}
\thanks{$^{2}$Author is with the Constructor Knowledge Labs (CKL), Bremen, Germany.}}%

\begin{document}

\maketitle
\thispagestyle{empty}
\pagestyle{empty}

\begin{abstract}
We promote in this paper the processing of radar data in the frequency domain to achieve higher robustness against noise and structural errors, especially in comparison to feature-based methods. This holds also for high dynamics in the scene, i.e., ego-motion of the vehicle with the sensor plus the presence of an unknown number of other moving objects. In addition to the high robustness, the processing in the frequency domain has the so far neglected advantage that the underlying correlation based methods used for, e.g., registration, provide information about all moving structures in the scene. 
A typical automotive application case is overtaking maneuvers, which in the context of autonomous racing are used here as a motivating example. Initial experiments and results with Fourier SOFT in 2D (FS2D) are presented that use the Boreas dataset to demonstrate radar-only-odometry, i.e., radar-odometry without sensor-fusion, to support our arguments.


\end{abstract}

\section{Introduction}
An important advantage of automotive radar 
is its robustness in challenging weather conditions \cite{automotive-sensors-adverseweather-survey-ISPRS23}, i.e., it is not as strongly affected by rain and snow as vision or lidar.
Also, it is relatively low-cost compared to lidar \cite{automotive-radar-lidar-survey-Chips23}. 
But radar delivers relatively low resolution and noisy data \cite{automotive-radar-lidar-survey-Chips23,automotive-radar-signalprocessing-survey-SPMag19}. Furthermore, it inherently suffers from interferences, which can be mitigated, 
 but which still is a factor in practice - up to the generation of ghost objects.

The use of feature for radar processing has hence its limitations - no matter whether they are engineered or generated by data-driven AI. The reason is that each feature only uses local information from a limited size kernel, which is easily corrupted by noise. When operating in the frequency domain, information on different scales is used ranging from small scale, local data up to larger structures. This leads in general to a higher robustness in the processing, e.g., for registration. In the context of registration, there is an other, so far neglected advantage.      

\begin{figure*}[htb]
\begin{center}
\includegraphics[width=.8\linewidth]{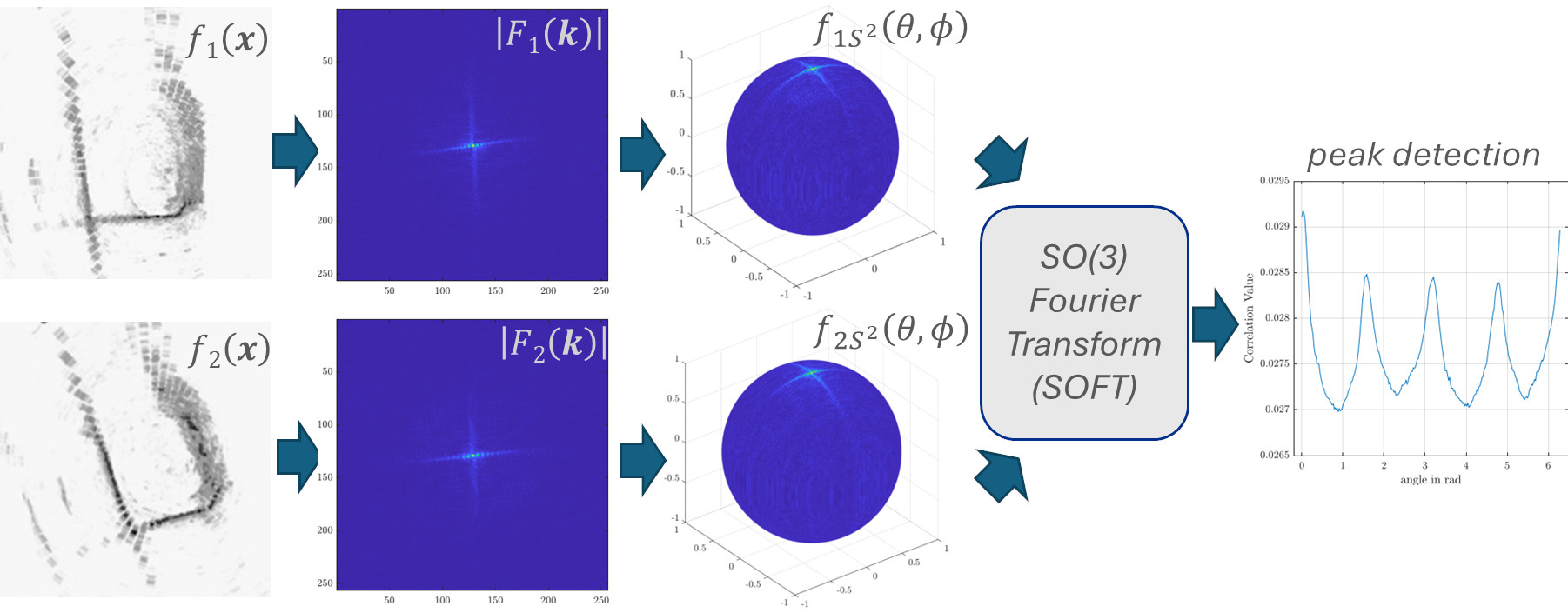}
\caption{Registration in the frequency domain, here of sonar data with FS2D, can not only deal very well with noisy data. There is also the so far neglected advantage that the underlying processing with correlation methods leads to peaks that indicate the transformation parameters for multiple moving structures in the data in parallel. \label{fig:spectral-registration-peaks}}
\end{center}
\end{figure*}

No matter whether, e.g., Fourier-Mellin \cite{reddy-FFT-registration96,chen-FourierMellin-image-registration94}, Fourier-Mellin-SOFT \cite{Fourier-Mellin-SOFT-FMS-7DoFregistration-IJCV18}, or Fourier-SOFT in 2D (FS2D) \cite{FS2D-MSSregistration-ICRA23} are used for the processing in the frequency domain, the underlying correlation methods inherently provide information about all moving structures in the scene. 

More precisely, they generate in theory a Dirac pulse for each degree-of-freedom (dof) for a rigid motion. These are in practice broader peaks due to the imperfections in the to be registered scans, i.e., noise, occlusions, partial overlap, etc. (Fig.\ref{fig:spectral-registration-peaks}). The peaks indicate the relative motion-parameters of the sensor-poses, i.e., they generate ego-motion estimates. In addition, a peak is generated within each dof for each moving structure in the sensor data, i.e., for each of the to be tracked objects. So, the motion of the objects is indicated by peaks together with the ego-motion. A major advantage of the processing in the frequency domain is hence that the number of dynamic objects does not need to be known a priori and that no involved methods are required to detect the objects.

The rest of this paper is structured as follows. Sec.\ref{sec:racing} provides a motivating application example in the context of autonomous overtaking maneuvers, especially in autonomous racing. Initial experiments and
results with Fourier SOFT in 2D (FS2D) are presented in Sec.\ref{sec:experiments-results} that demonstrate radar-only-odometry, i.e., radar-odometry without sensor-fusion, to support our arguments. 
Sec.\ref{sec:conclusion} concludes the paper.


\section{Dynamics in the Scene as Challenge for the Registration of Automotive Radar}
\label{sec:racing}

\begin{figure}[!htb]
\begin{center}
\includegraphics[width=\linewidth]{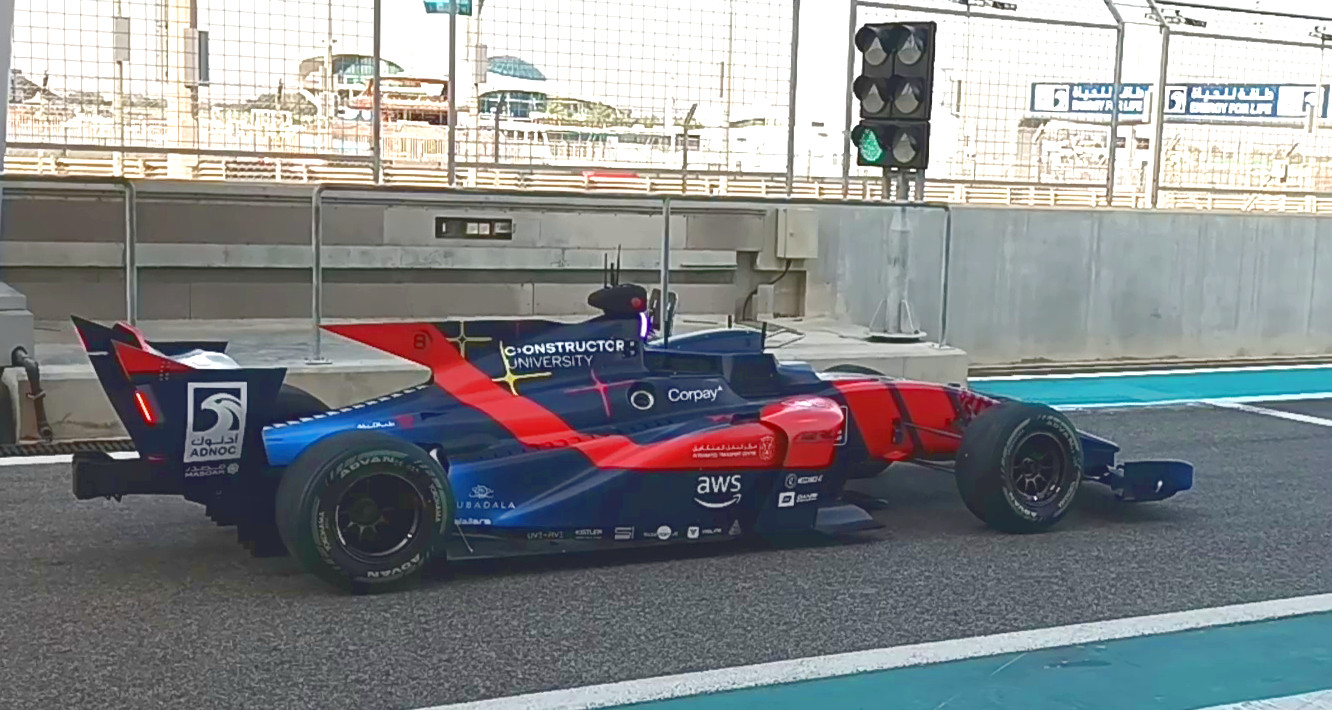}
\caption{The Constructor University (CU) vehicle used for high-velocity autonomous driving. \label{fig:A2RL-ConstructorCar-01}}
\end{center}
\end{figure}

\begin{figure}[!htb]
\begin{center}
\includegraphics[width=\linewidth]{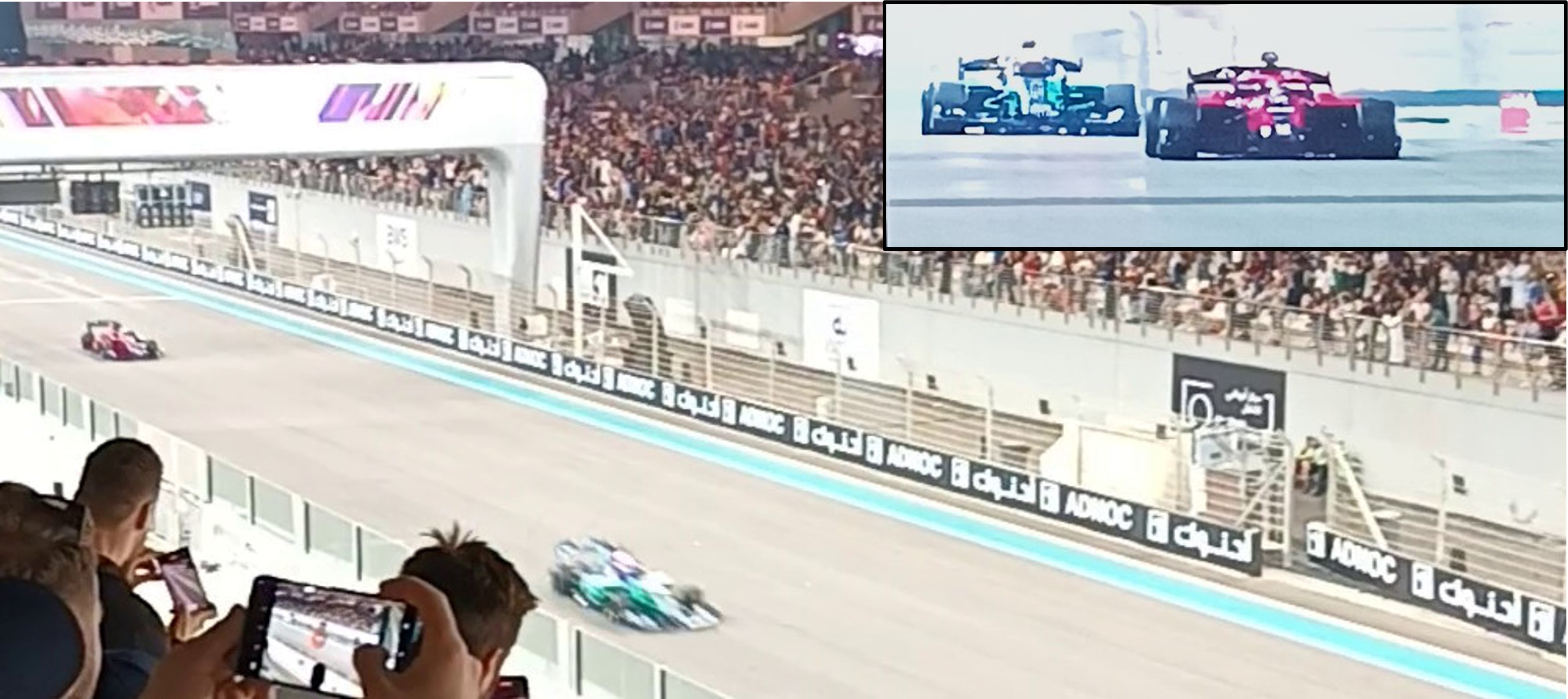}
\caption{The A2RL races feature fully autonomous driving at high velocities on the Yas Marina Formula-1 race-track in Abu Dhabi. \label{fig:overtaking-A2RL}}
\end{center}
\end{figure}

\begin{figure}[!htb]
\begin{center}
\includegraphics[width=\linewidth]{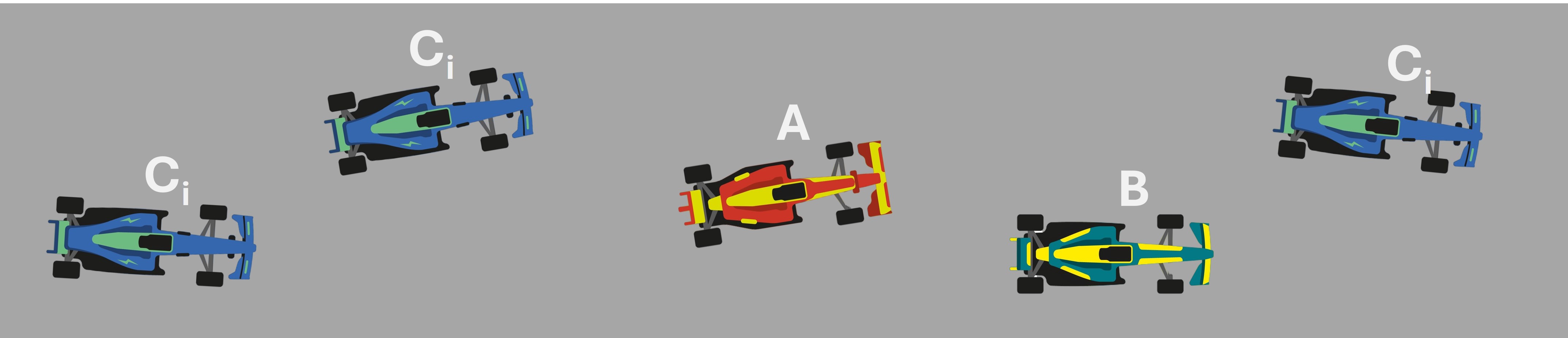}
\caption{Autonomous overtaking 
includes the perception challenge of Multi-Object-Tracking (MOT). For planning and executing the maneuver, the vehicle $A$ itself, the car $B$ that is to be overtaken, as well as an unknown number of vehicles $C_i$ in the vicinity ($i \geq 0$) have to be tracked. 
\label{fig:overtaking-illustration}}
\end{center}
\end{figure}

Autonomous overtaking \cite{automotive-overtaking-survey-taxonomy-VC23} is a very challenging maneuver in the context of  autonomous driving, respectively higher-level Advanced Driver Assistance Systems (ADAS) \cite{automotive-autonomous-driving-survey-ExpSysAppl24,automotive-autonomous-driving-methods-survey-Access20,automotive-autonomous-driving-survey-ExpSysAppl20}. This holds especially for high-velocity driving (Fig.~\ref{fig:A2RL-ConstructorCar-01} and \ref{fig:overtaking-A2RL}) that pushes methods to their limits with respect to the requirements in terms of real-time performance and robustness
\cite{automotive-racing-survey-JITS22,automotive-racing-learning-MCS18}. 

Note that two dynamic objects have to be at least tracked in the context of overtaking maneuvers (Fig.~\ref{fig:overtaking-illustration}), namely the moving platform $A$ itself and the vehicle $B$ that is to be overtaken. Furthermore, there may be one or more vehicles next to or behind $A$ as well as one or more additional vehicles next to or in front of $B$, which all can play an important role in the decision whether, respectively how to overtake $B$, i.e., which have to be precisely tracked, too. Last but not least, general traffic scenarios may involve additional dynamic objects like pedestrians, bicycles, etc.

Multi-Object-Tracking (MOT) in general is a very well established research field, especially in Computer Vision when using conventional imaging sensors, i.e., cameras \cite{mot-multi-object-tracking-survey-AI21}. But this does not hold with respect to the use of radar - though there is a clear increase in work on radar-based motion estimation in recent time \cite{automotive-radar-odometry-survey-ITITS24,automotive-MOT-4Dradar-comparison-IV24}. MOT with vision tends to rely on features, i.e., very local information, which already tend to fail for ego-motion estimation due to the - in comparison to cameras - poor quality of radar data \cite{automotive-radar-odometry-limits-Sensors23}. Motion-estimation with radar hence tends to use more robust alternatives like the Normal Distribution Transform (NDT) \cite{automotive-radar-odometry-NDT-RAS17} or Gaussian Mixture Models (GMM) \cite{automotive-radar-odometry-GMM-RAL22} as well as classical methods operating in the frequency domain like the Fourier-Mellin-Transform \cite{automotive-radar-registration-FourierMellin-ICRA20}.  

The foundations of using the frequency domain for image processing have been laid more than half a century ago \cite{FFT-ImageRegistration-ITGE70}. 
But there are still conceptual advancements; for example, the extension of Fourier-Mellin \cite{chen-FourierMellin-image-registration94,reddy-FFT-registration96} to 3D \cite{Fourier-Mellin-SOFT-FMS-7DoFregistration-IJCV18}. To illustrate the potential of operations in the frequency domain, we use here own work in form of Fourier-SOFT in 2D (FS2D) \cite{FS2D-MSSregistration-ICRA23}. 

\section{Radar Odometry without Sensor Fusion}
\label{sec:experiments-results}

A detailed description of FS2D can be found in \cite{FS2D-MSSregistration-ICRA23} where it is introduced and used for the registration of underwater sonar, more precisely, of data from a low-cost mechanically scanning sonar (MSS) on a low-cost unmanned underwater vehicle (UUV). An important element of FS2D is its use of the SO(3) Fourier transform (SOFT) \cite{FFTsOnTheRotationGroup-Journal08} to estimate the 2D rotation based on a projection of the spectral magnitudes of the frequency domain representation of the data onto a SO(3) sphere.
  

\begin{figure}[!h]
    \centering
    \includegraphics[width=\linewidth]{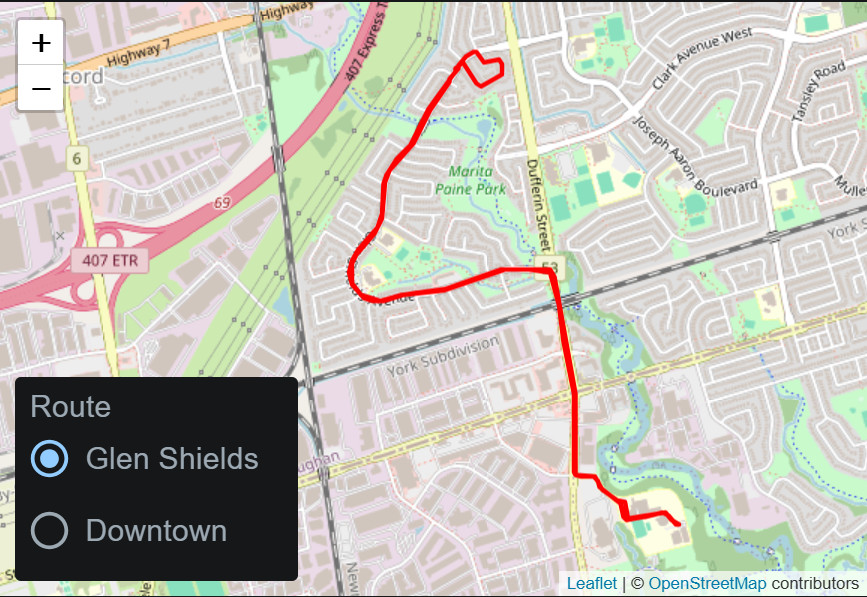}
    \caption{The Glen Shield route from the Boreas dataset (image from the Boreas website \url{https://www.boreas.utias.utoronto.ca/} using OpenStreetMap).}
    \label{fig:trajectoryOpenStreetMap}
\end{figure}


\begin{figure}[!h]\vspace{+5pt}
    \centering
    \begin{subfigure}{.7\linewidth}
        \includegraphics[width=\linewidth]{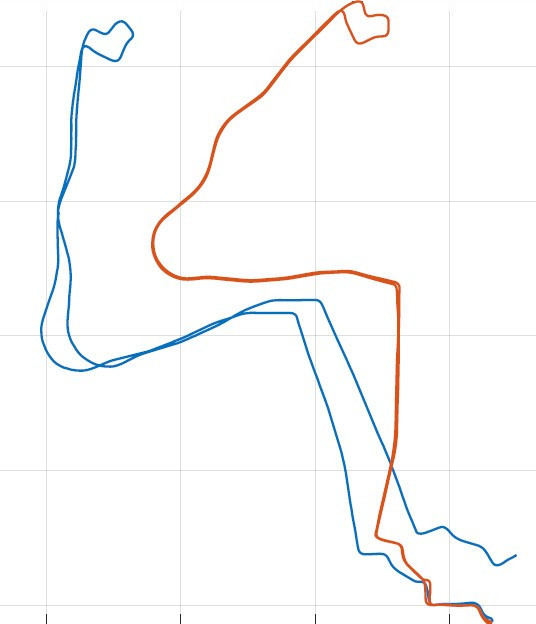}
        \caption{2020-11-26-13-58 \label{fig:trajectories-a}}
    \end{subfigure}
    \hspace{2em}
    \begin{subfigure}{.7\linewidth}
        \includegraphics[width=\linewidth]{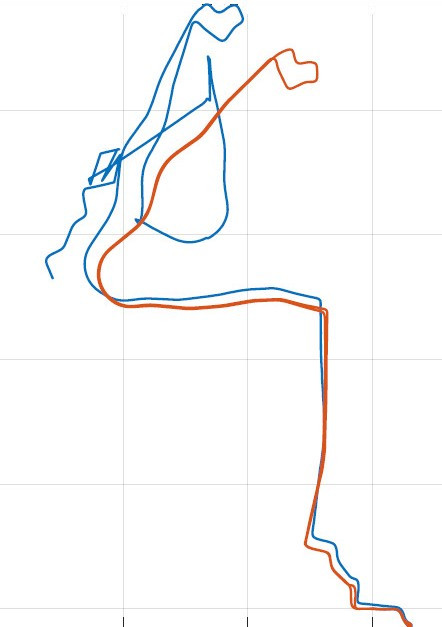}
        \caption{2020-12-01-13-26 \label{fig:trajectories-b}}
    \end{subfigure}
    \caption{The estimated trajectories (blue) and the ground truth (red) for the first two datasets as illustrative examples. Note that the estimated trajectories are purely based on pairwise registration of radar scans without any other localization or motion estimation, and that they include all outliers.}
    \label{fig:trajectories}
\end{figure}

\begin{figure*}[!h]\vspace{+5pt}
    \centering
        \includegraphics[width=.31\linewidth]{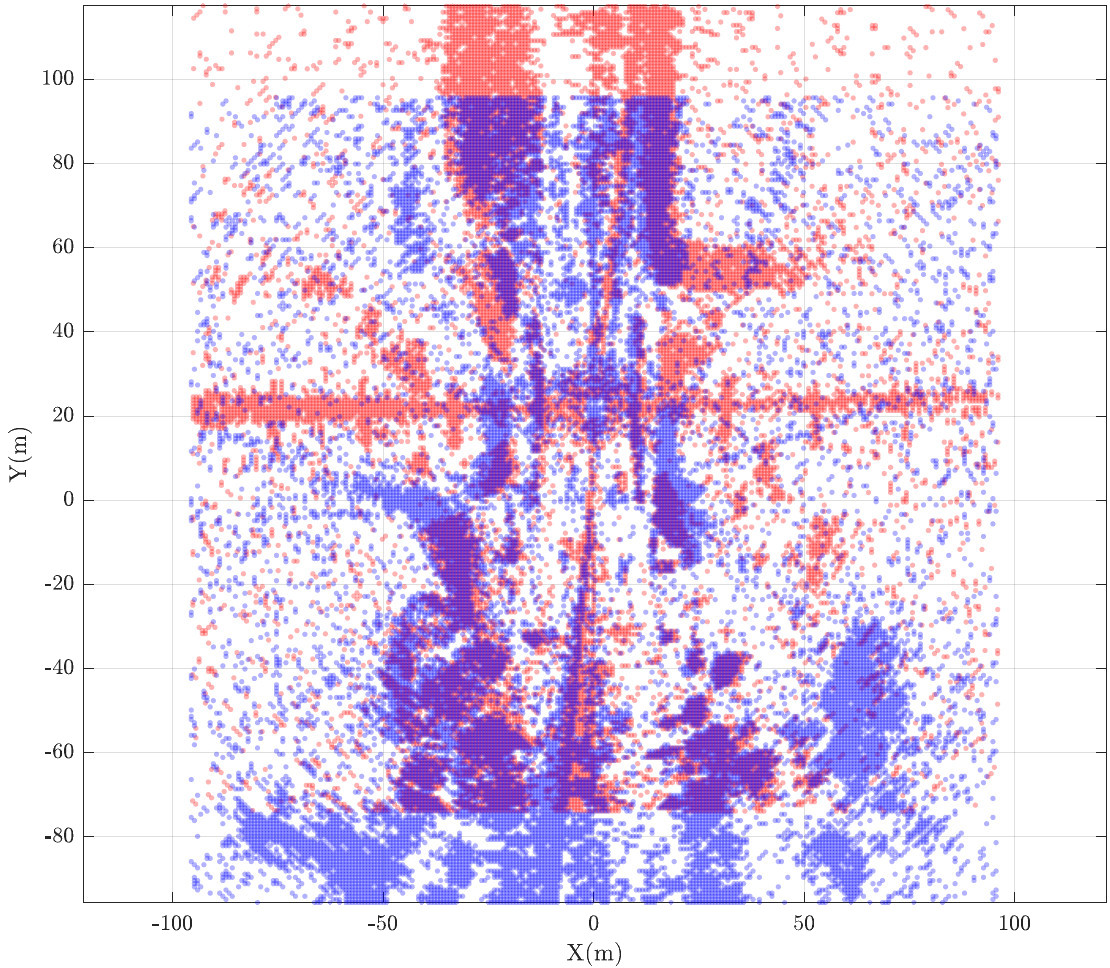}
        \includegraphics[width=.31\linewidth]{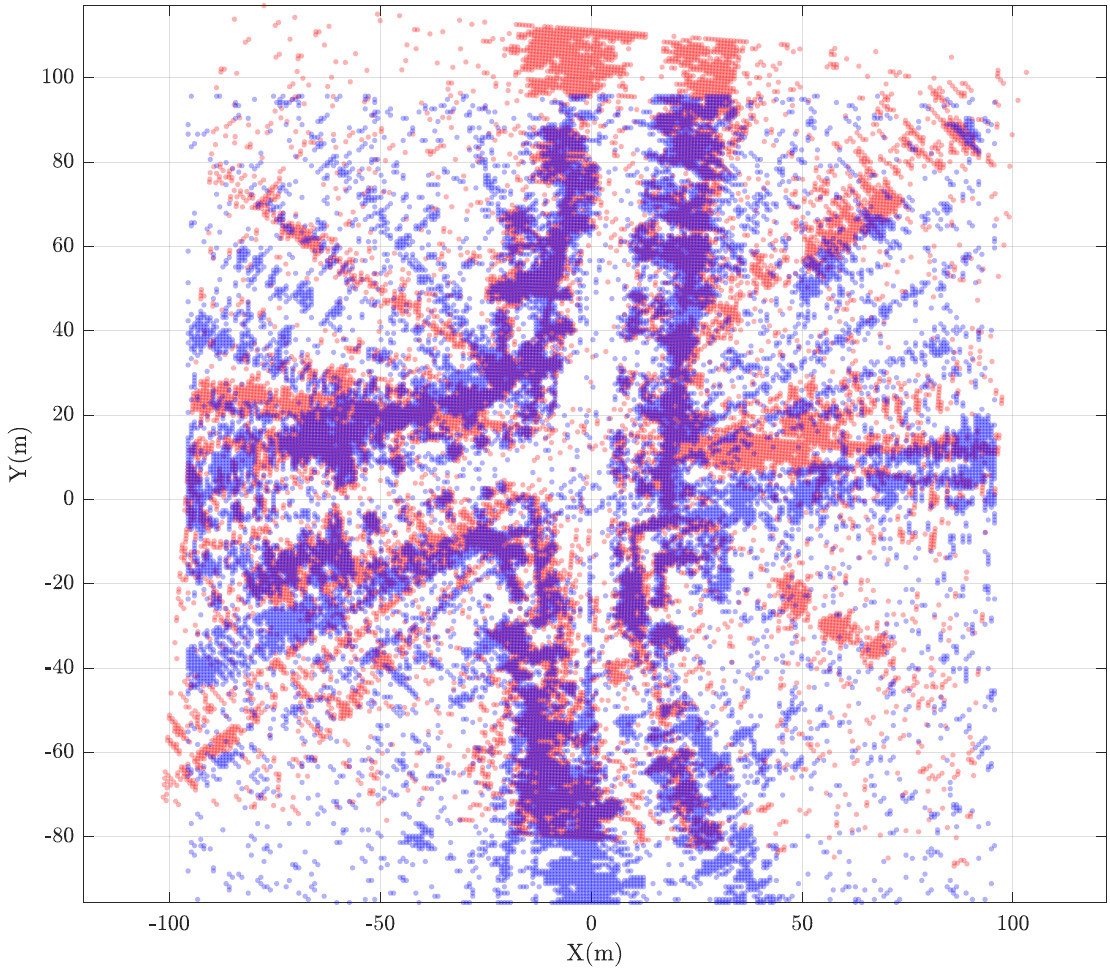}
        \includegraphics[width=.31\linewidth]{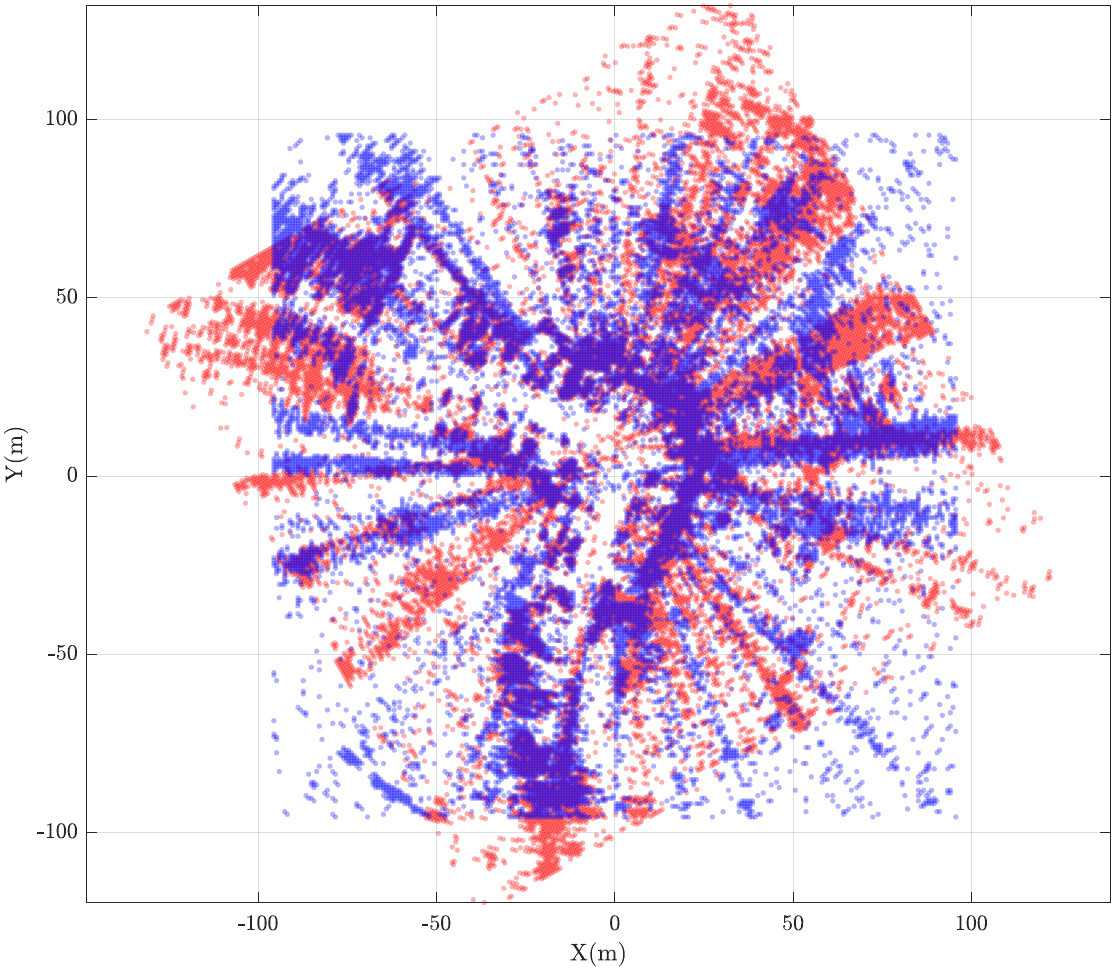}
			  \includegraphics[width=.31\linewidth]{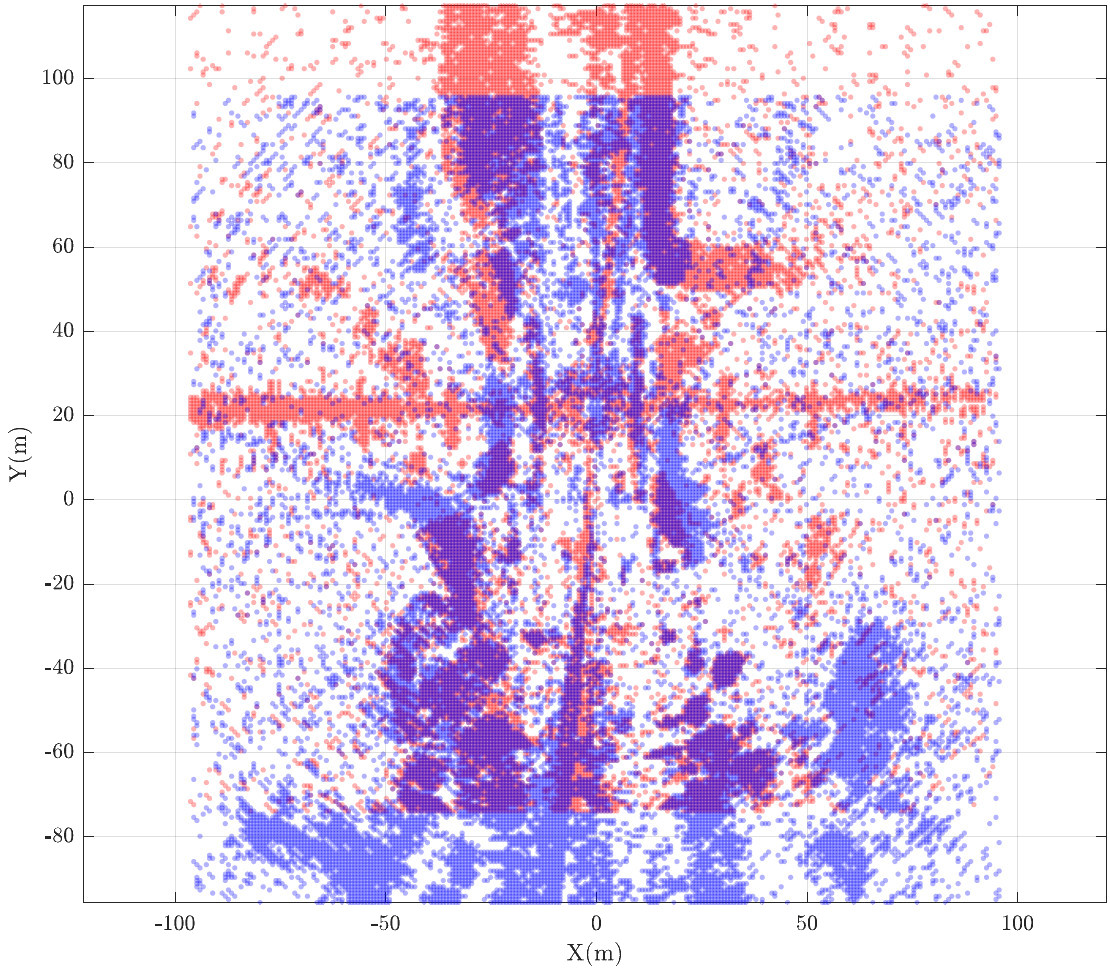}
        \includegraphics[width=.31\linewidth]{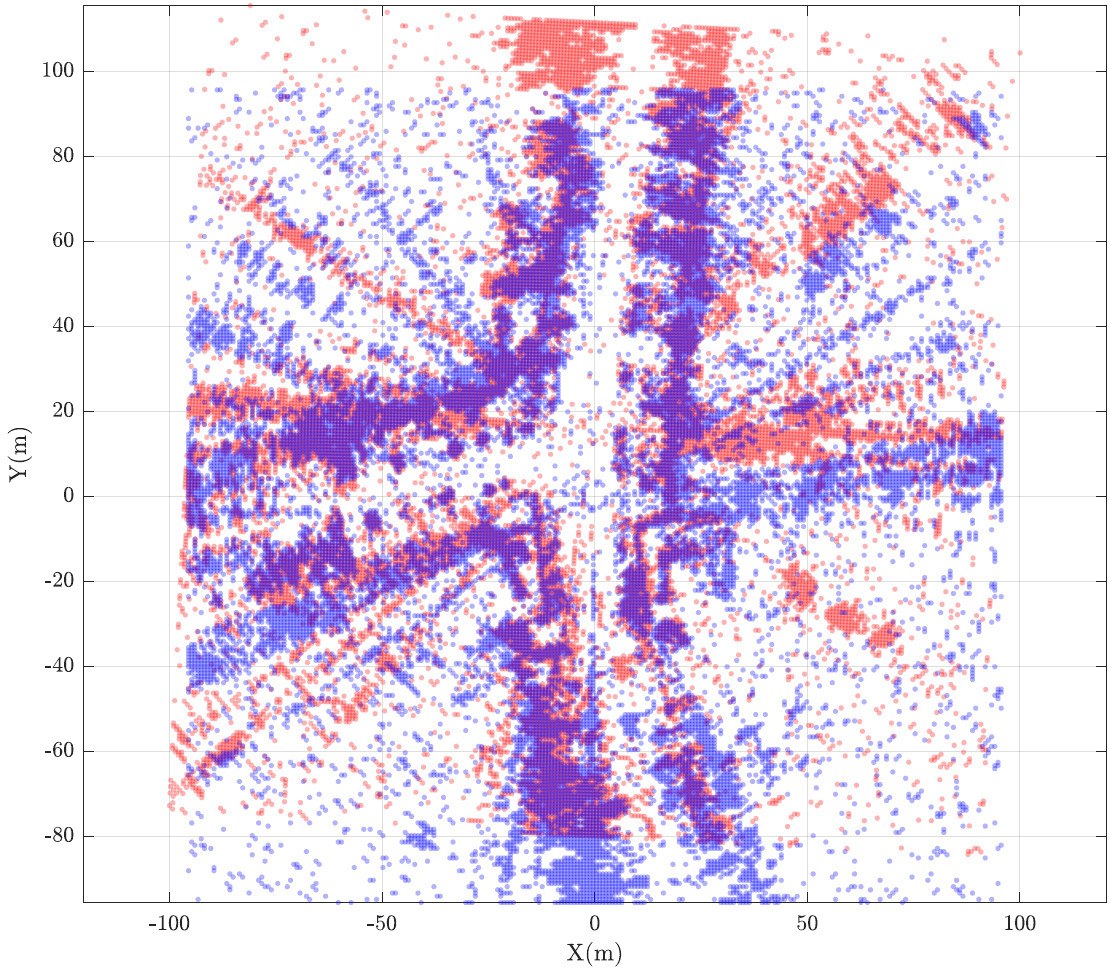}
        \includegraphics[width=.31\linewidth]{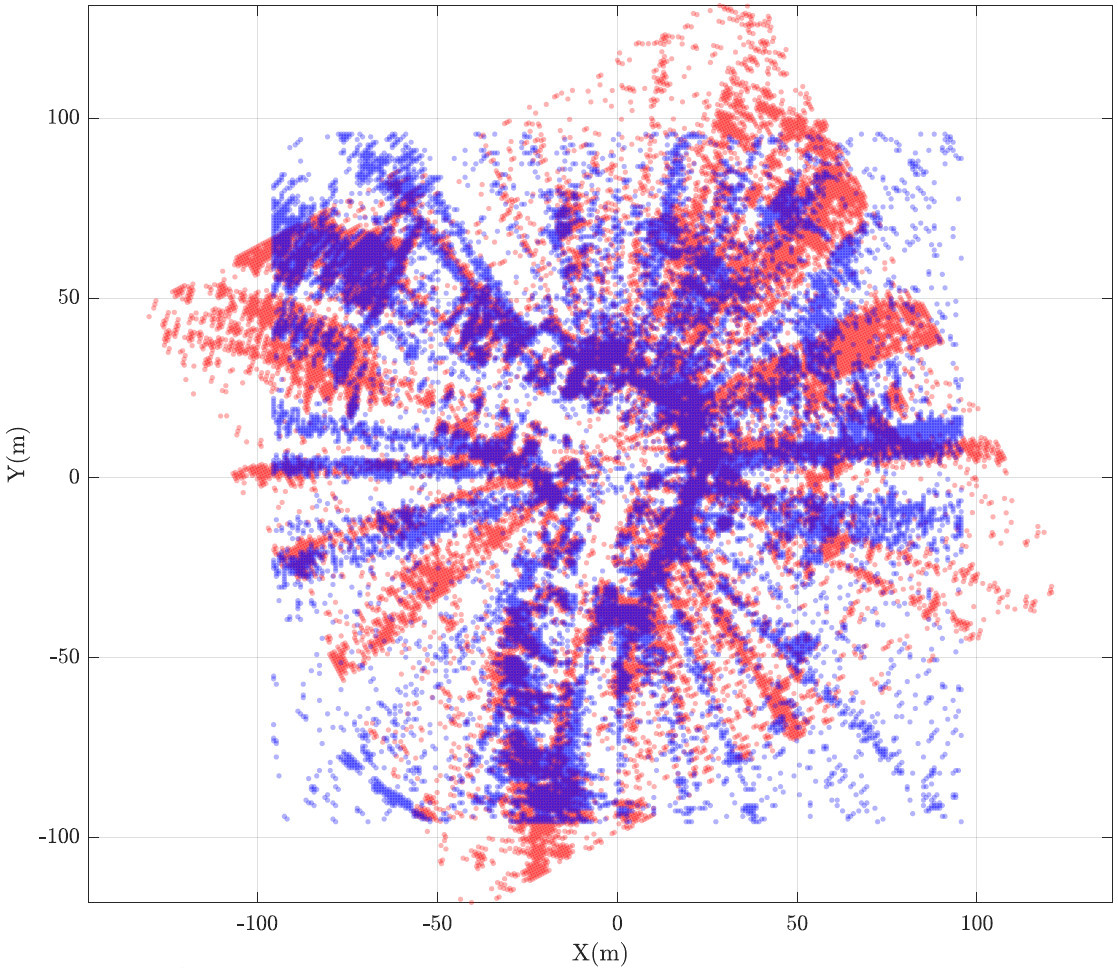}
    \caption{Three examples of the FS2D registration. On the top, an overlay of two scans in red and blue is shown using their ground truth poses. The actual motion is hence represented in the translation and rotation of the blue scan relative to the red one. Below, the same scan pairs are projected according to the motion estimates from FS2D, which are very close to the ground truth.}
    \label{fig:pclMatching}
\end{figure*}



To illustrate the potential of FS2D as registration method in the frequency domain, we show that it can be used for radar odomotry without sensor fusion. The Boreas dataset \cite{AutonomousDriving-BoreasDataSet-IJRR23} with a Navtech CIR304-H radar is used to this end. The motion estimates are generated by solely using pairwise registrations with FS2D - in contrast to the the state-of-the-art where the fusion of multiple sensor modalities is the standard 
\cite{automotive-localization-survey-Access23,automotive-localization-survey-performance-Access23,automotive-localization-survey-Sensors22,automotive-localization-survey-ITJ18}.
%
%

The data is far from perfect (see also Fig.\ref{fig:pclMatching}) as the Navtech radar suffers from both motion distortion and Doppler distortion. Both effects can be compensated and the state-of-the-art considers this to be important \cite{Radar-MotionCompensation-DopplerCompensation-RAL21,RadarOdometry-RSS21}. This is understandable as  features, no matter whether they are engineered or generated by data-drivn AI, rely on local information. FS2D takes  so to say information on all scale levels into account as it operates in the frequncy domain. It is hence much less vulnerable to these structural error sources.
%
%
As illustrated in Fig.\ref{fig:pclMatching}, additional error sources include high amounts of salt and pepper noise, ghost beams, and especially dynamics created by moving objects, e.g., other cars. The results for scenes with high dynamics indicate that FS2D copes very well with these situations. 

 
For our experiments, we use all 32 sequences in the data-set that come with ground truth localization, which are along the Glen Field route (Fig.\ref{fig:trajectoryOpenStreetMap}). 
We use only every 5th scan, i.e., scan number $i$ is registered with scan number $i+5$, as this is already sufficient for good motion estimates while it allows real-time processing already with the current plain vanilla implementation of FS2D on a single core of a standard CPU.

Three illustrative examples of single registration results are shown in Fig.\ref{fig:pclMatching}, i.e., three pairs of a scan $A$ (blue) and a scan $B$ (red) that are projected using their ground truth poses (top) and according to the FS2D results (bottom).  First of all, the images illustrate the high amounts of noise and structural errors in the input data. Second, they show that despite these challenges, FS2D provides motion estimates that are congruent with ground truth. 

Fig.\ref{fig:trajectories} show two full trajectories as examples, i.e., the motion estimates derived from full sequences of registrations using the first two recordings in the dataset. These are the "2020-11-26-13-58" recording shown on the left, and the "2020-12-01-13-26" recording shown on the right, with each of the ground truth trajectories shown in red. The trajectories of the 827, respectively 768 sequential poses estimated with FS2D are shown in blue. The two examples include outliers and experience some drift, which is to be expected as there are no global corrections. Nonetheless, the examples illustrate that there is an overall very reasonable odometry estimate purely based on radar and despite some dynamics in the scene.
 


This is also supported by a quantitative analysis using all 32 sequences with ground truth. In total, 26,598 pairs of radar data are registered. An uncertainty estimation of the registration result allows detecting outliers \cite{underwater-vSLAM-icra10}, which are rejected. The estimation of the rotation, which tends to be the hard part for registration, has an average error of 0.62$^o$ with only 1.19\% outliers. The error in the translation estimation may look quite large at first glance with an average of 0.49\,{\em m}. But with a cell size of $0.75m \times 0.75m$, the expected average error due to this discretization in the translation estimation is already $\sqrt{2}/2\cdot 0.75 m = 0.53 m$, which is a fixed offset in the average translation errors. There are various ways to efficiently cope with this in the signal processing, which are left for future work. 

\section{Conclusions}
\label{sec:conclusion}

We made a case in this paper that processing in the frequency domain has a high potential for handling automotive radar data. To this end, initial experiments and results with Fourier SOFT in 2D (FS2D) as a new registration method are presented that demonstrate radar-only-odometry, i.e., radar-odometry without sensor-fusion. Furthermore, we pointed to the so far neglected aspect that in the frequency domain, the underlying correlation methods inherently provide information about all moving structures in the scene. This not only increases robustness in dynamic scenes, but it can also substantially ease multi-object-tracking on noisy data.

\end{document}